%% file: main.tex
\documentclass[sigconf,nonacm]{acmart}
\usepackage{rotating}
\usepackage{multirow}

\AtBeginDocument{%
  \providecommand\BibTeX{{%
    \normalfont B\kern-0.5em{\scshape i\kern-0.25em b}\kern-0.8em\TeX}}}

\setcopyright{acmcopyright}
\copyrightyear{2018}
\acmYear{2018}
\acmDOI{XXXXXXX.XXXXXXX}

\acmPrice{}
\acmISBN{}

\begin{document}

\title{Leveraging Pre-trained Language Models for Time Interval Prediction in Text-Enhanced Temporal Knowledge Graphs}
\author{Duygu Sezen Islakoglu}
\authornote{Contact Author}
\affiliation{\institution{Utrecht University}\country{Netherlands}}
\email{d.s.islakoglu@uu.nl}

\author{Mel Chekol}
\affiliation{\institution{Utrecht University}\country{Netherlands}}
\email{m.w.chekol@uu.nl}

\author{Yannis Velegrakis}
\affiliation{\institution{Utrecht University}\country{Netherlands}}
\email{i.velegrakis@uu.nl}

\renewcommand{\shortauthors}{Islakoglu, et al.}

\begin{abstract}
Most knowledge graph completion (KGC) methods learn latent representations of entities and relations of a given graph by mapping them into a vector space. Although the majority of these methods focus on static knowledge graphs, a large number of publicly available KGs contain temporal information stating the time instant/period over which a certain fact has been true. Such graphs are often known as temporal knowledge graphs. Furthermore, knowledge graphs may also contain textual descriptions of entities and relations. Both temporal information and textual descriptions are not taken into account during representation learning by static KGC methods, and only structural information of the graph is leveraged. Recently, some studies have used temporal information to improve link prediction, yet they do not exploit textual descriptions and do not support inductive inference (prediction on entities that have not been seen in training).
 
We propose a novel framework called TEMT that exploits the power of pre-trained language models (PLMs) for text-enhanced temporal knowledge graph completion. The knowledge stored in the parameters of a PLM allows TEMT to produce rich semantic representations of facts and to generalize on previously unseen entities. TEMT leverages textual and temporal information available in a KG, treats them separately, and fuses them to get plausibility scores of facts. Unlike previous approaches, TEMT effectively captures dependencies across different time points and enables predictions on unseen entities. To assess the performance of TEMT, we carried out several experiments including time interval prediction, both in transductive and inductive settings, and triple classification. The experimental results show that TEMT is competitive with the state-of-the-art.
\end{abstract}

\maketitle

\input{introduction.tex}
\input{preliminaries.tex}

\input{proposed-framework.tex}
\input{experiments.tex}

\input{related-work.tex}
\input{conclusion.tex}

\bibliographystyle{ACM-Reference-Format}
\bibliography{references}

\appendix
\input{appendix}

\end{document}

%% file: introduction.tex
\section{Introduction} \label{introduction}
Knowledge Graphs (KGs) are directed graphs that model real-world entities and relationships as nodes and edges between them. 
They store knowledge, i.e., facts, in a structured and interpretable way, typically as triples of the form $\langle$subject, relation, object$\rangle$, e.g. $\langle$Obama, presidentOf, U.S.$\rangle$. They find a wide range of applications and have been used in recommendation systems, question-answering systems, and many knowledge-intensive systems, including virtual assistants and chatbots \citep{https://doi.org/10.48550/arxiv.2009.11564}.

Knowledge graphs are often incomplete, meaning that some elements of the facts are not available. To this end, KG completion (KGC) methods aim at finding missing links between the entities. Most of the studies on KG completion have focused on static knowledge graphs where the graph remains unchanged over time~\cite{Ji_2022}. However, in real life, facts are not always valid throughout time, but only in specific time periods. 
For instance, presidents of a country are valid only throughout their term. To model this, triples in a KG may have validity time intervals associated with them. This additional information converts a triple into a quadruple form $\langle$subject, relation, object, time interval$\rangle$, e.g. $\langle$Obama, presidentOf, U.S., [2009, 2017]$\rangle$. A graph consisting of a set of such temporal facts -quadruples- is referred to as a \textit{Temporal Knowledge Graph} (TKG).  

\begin{figure}[t!]
    \centering
    \captionsetup{justification=centering,margin=1cm}
    \includegraphics[width=0.43\textwidth]{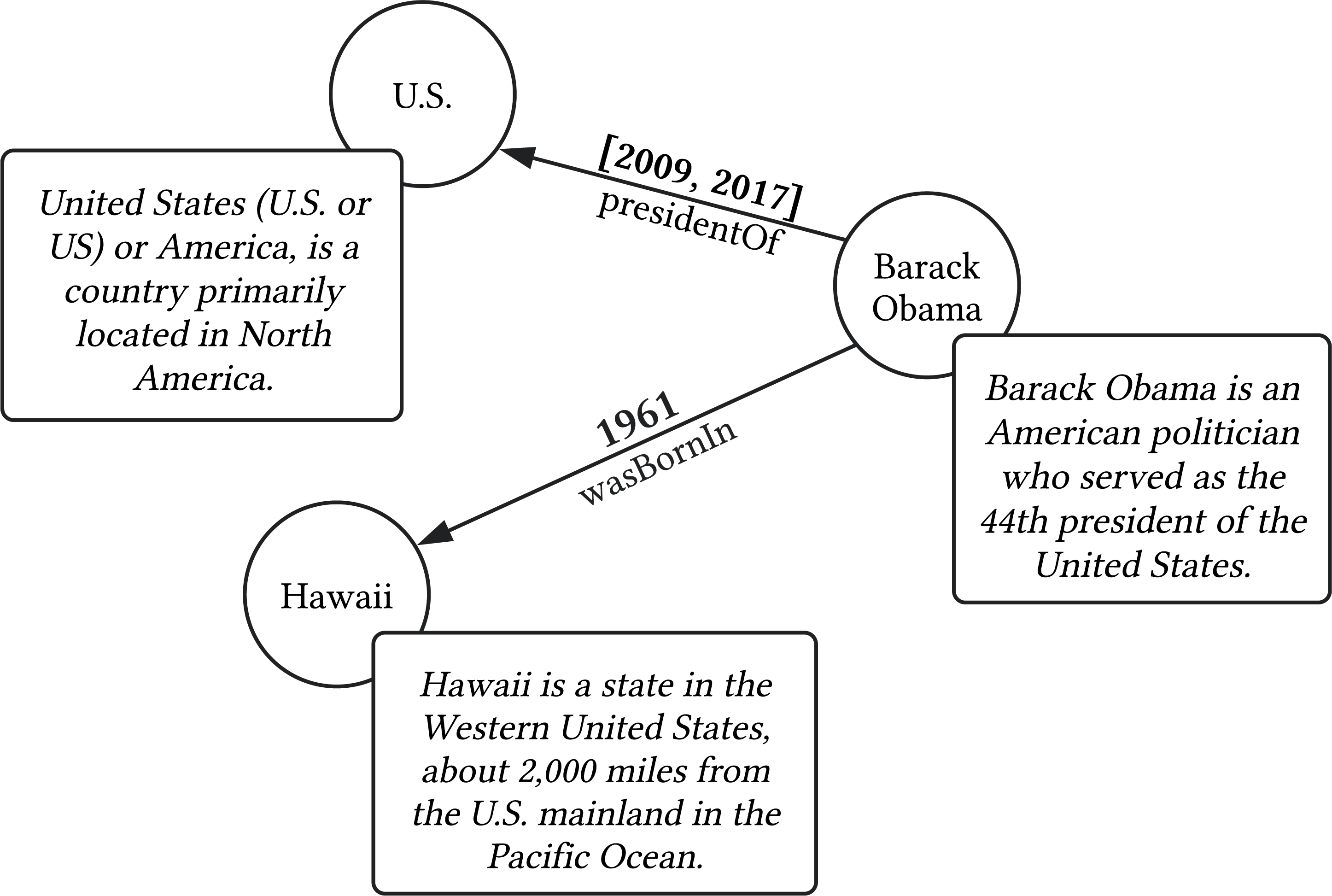}
        \caption{Text-Enhanced Temporal Knowledge Graph}
    \label{Text-Enhanced-Temporal-Knowledge-Graph} 
    \end{figure}
    
Alike traditional KGs, temporal knowledge graphs suffer inherently from incompleteness due to their dynamic nature. In particular, the temporal information (validity time interval) is often missing after an automatic knowledge graph construction. To this end, the temporal knowledge graph completion (TKGC) task aims to predict a missing quadruple element, such as time interval prediction $\langle$s, r, o, ?$\rangle$. The time interval prediction task is useful for temporal question answering, automatic construction, and verification of temporal knowledge graphs that have temporal constraints~\citep{https://doi.org/10.48550/arxiv.2009.11564}. 

Textual descriptions of entities and relations can significantly enhance the effectiveness of (temporal) knowledge graphs completion methods~\cite{xie_representation_2016} since they contain valuable information regarding semantic relationships across entities. For instance, in a description of an entity there may be a reference to some other entity, despite the absence of any type of relationship (edge) in the knowledge graph among them. Unfortunately, most of the existing works on temporal knowledge graph completion do not fully exploit this additional information in downstream tasks. Furthermore, by considering the semantics of the descriptions, one may gain insight into the validity time of the facts. 
For instance, if an entity description contains elements from a certain century or a period like Renaissance, facts involving that entity may be valid for that period. Figure~\ref{Text-Enhanced-Temporal-Knowledge-Graph} illustrates a temporal knowledge graph with entities that have textual descriptions associated with them. The fact that Obama has been the 44th president, may increase the plausibility that his presidency started in 2009. 

The availability of textual descriptions in knowledge graphs provides an excellent opportunity for exploiting the benefits that both knowledge graphs and language models can offer. Recent works has shown that language models store real-world knowledge in their parameters and can potentially be used as knowledge graphs~\citep{petroni-etal-2019-language, https://doi.org/10.48550/arxiv.2204.06031}, and that textual information can improve link prediction for static knowledge graphs~\citep{li_siamese_2021, yao_kg-bert_2019}. Furthermore, it has been shown that entity descriptions and pre-trained language models can model facts that involve unseen entities (inductiveness)~\citep{daza_inductive_2021, wang_simkgc_2022}.
The ability to do inductive reasoning is crucial since most real-world knowledge graphs are often continuously extended with new entities. 
Unfortunately, most temporal knowledge graph completion mechanisms are transductive, which means that they can only perform predictions on the entities they have already seen during training, i.e., are part of their training set. 

We provide a novel method for temporal knowledge graph completion that leverages the available textual and temporal information for time interval prediction. This is achieved by encoding text and time separately and fusing them to predict the plausibility score of a quadruple.
As text encoder, we use a pre-trained language model to get a contextualized triple embedding and to generalize on unseen entities. 
We have implemented this idea in a framework called TEMT (\textbf{T}ext \textbf{E}ncoder \textbf{M}eets \textbf{T}ime)\footnote{The datasets and the source code are available at \url{https://github.com/duyguislakoglu/TEMT}} for time interval prediction (Section~\ref{sec:framework}). We enhance the two real-world temporal knowledge graphs YAGO11k and Wikidata12k~\citep{dasgupta-etal-2018-hyte} with textual names and entity descriptions and generate inductive splits. Using these datasets, we carry out experiments on transductive and inductive time interval prediction tasks (Section~\ref{sec:experiments}). The results show that TEMT is competitive with state-of-the-art approaches. 
In particular, the inductive time interval prediction results show that TEMT is able to reason on unseen entities (even when both the subject and object entities of a quadruple are unseen in training). 

%% file: preliminaries.tex
\begin{figure*}[h!]
    \centering
    \captionsetup{justification=centering,margin=1cm}
    \includegraphics[width=0.58\textwidth]{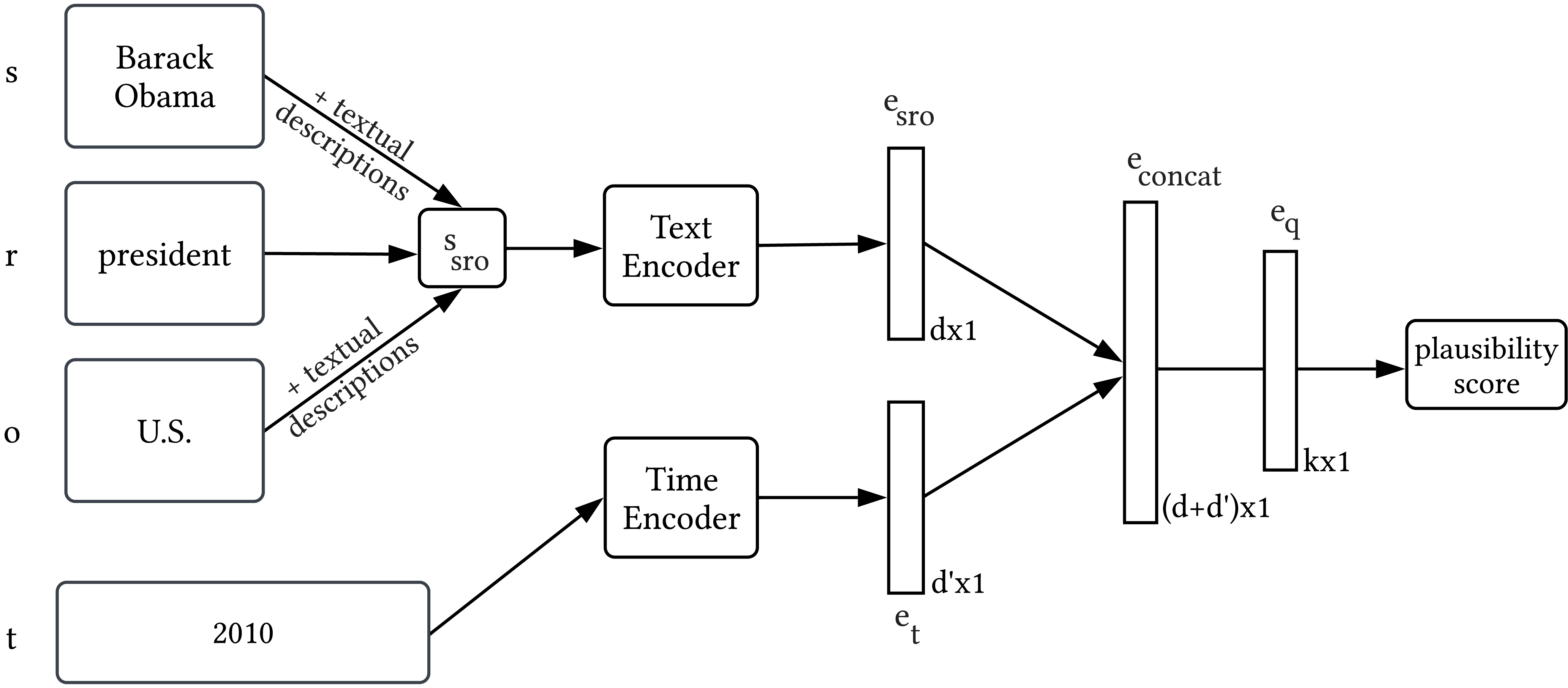}
        \caption{TEMT Framework}
    \label{Framework} 
    \end{figure*}
    
\section{Problem Statement} \label{preliminaries}
A temporal knowledge graph (TKG) is a directed graph $\mathcal{G} = (\mathcal{E}, \mathcal{R}, \mathcal{T}, \\\mathcal{Q})$ where $\mathcal{E}, \mathcal{R}, \mathcal{T}$ are sets of entities, relations, time points and $\mathcal{Q}$ represents the set of quadruples (or temporal facts) in the format $\langle$subject $s$, relation $r$, object $o$, time interval $t_I$$\rangle$ where $s, o \in \mathcal{E}$, $r \in \mathcal{R}$, $t_I = [t_{start}, t_{end}]$ and $t_{start}$, $t_{end}\in \mathcal{T}$. Note that $t_I$ can also be a single time point if $t_{start}$=$t_{end}$.

Temporal knowledge graphs can be grouped into two types: event TKGs and interval-based TKGs. The former refers to TKGs in which $t_{start}=t_{end}$ for every time interval of every quadruple, i.e., each triple $\langle$s, r, o$\rangle$ in the graph is valid at a single time point. The format of time points depends on the chosen time granularity, such as years, months, or days. On the other hand, interval-based TKGs, which are the focus of this paper, represent TKGs where each quadruple has a validity time interval $t_I=[t_{start}, t_{end}]$. An interval is called left-open if $t_{start}$ is unknown, right-open if $t_{end}$ is unknown, and closed interval if both the start and end points are known.

Text-enhanced temporal knowledge graphs are TKGs where each entity and relation is associated with a name and each entity has some natural language text that describes its meaning.  This additional information provides a context and attaches a semantic meaning to the facts, which can be informative for predicting the validity time interval of the fact. An example of a text-enhanced temporal knowledge graph is given in Figure~\ref{Text-Enhanced-Temporal-Knowledge-Graph}. More formally, a text-enhanced temporal knowledge graph is a directed graph $\mathcal{G} = (\mathcal{E}, \mathcal{R}, \mathcal{T}, \mathcal{Q}, \mathcal{N}, \mathcal{D})$ where $\mathcal{E}, \mathcal{R}, \mathcal{T}$ are sets of entities, relations, time points and $\mathcal{Q}$ represents the set of quadruples, $\mathcal{N}$ denotes the set of entity and relation names, and $\mathcal{D}$ denotes the set of entity descriptions. We can split $\mathcal{Q}$ into three disjoint sets as train, validation, and test sets. Formally, \begin{math} \mathcal{Q} = \mathcal{Q}_{train} \cup \mathcal{Q}_{val} \cup \mathcal{Q}_{test} \end{math} where $\mathcal{Q}_{train}$ represents the set of train quadruples, $\mathcal{Q}_{val}$ represents the set of validation quadruples and $\mathcal{Q}_{test}$ represents the set of test quadruples. Similarly, we can specify the set of entities as $\mathcal{E}_{train}$, $\mathcal{E}_{val}$ and $\mathcal{E}_{test}$.

Time interval prediction is the task of predicting the validity time interval of a triple. More formally, given a quadruple $\langle$s, r, o, ?$\rangle$ with unknown validity time interval, the objective is to predict a time interval $t_I$. We can reformulate the question as follows: given training and validation sets, $\mathcal{Q}_{train}$ and $\mathcal{Q}_{val}$, the time interval prediction is to output $t_I$ for each test quadruple $\langle$s, r, o, ?$\rangle$ $\in$ $\mathcal{Q}_{test}$.

In this paper, we focus on two variants of this problem:  \textit{transductive time interval prediction} and \textit{inductive time interval prediction}. Transductive time interval prediction aims to predict $t_I$ for each test quadruple in $\mathcal{Q}_{test}$ that does not contain any new entities, i.e., \begin{math} \mathcal{Q}_{test}= \{\langle s, r, o, ? \rangle | s, o \in \mathcal{E}_{train}\}\end{math}. In other words, all entities in the test set are included in the training set, i.e., $\mathcal{E}_{test} \subseteq \mathcal{E}_{train}$. Furthermore, inductive time interval prediction aims to predict the time interval of facts where the test set contains previously unseen entities. So for each quadruple in the test set, the subject or the object (or both) does not appear in $\mathcal{E}_{train}$. Therefore, the test quadruples \begin{math}
\mathcal{Q}_{test}= \{\langle s, r, o, ? \rangle | s  \not \in \mathcal{E}_{train}\} \cup \{\langle s, r, o, ? \rangle | o  \not \in \mathcal{E}_{train}\} 
\end{math}.

\section{Pre-trained language models}
We exploit the representational power of pre-trained language models to model the semantics of facts and to deal with unseen entities. 
Language models  assign a probability to a word by taking into account the other words in a sentence and can predict the next word given a sequence of words. This can be done by learning a latent representation of words in a vector space. Moreover, models such as bidirectional encoder representations from transformers (BERT) \citep{devlin-etal-2019-bert} do not only consider the previous words but also take the subsequent words into account. BERT generates a contextual word embedding where the representation of a word depends on the whole context. 

A pre-trained language model (PLMs) is a language model that is trained on a large text corpora including books, encyclopedias, and web data. PLMs can be used for many downstream tasks such as question-answering and text summarization. A pre-trained model, such as pre-trained BERT, can be further fine-tuned for a specific task or can be used for feature extraction of a sentence. However, since BERT is designed for word-level tasks and not optimized for sentence-level tasks, it performs poorly in semantic textual similarity tasks \citep{reimers-2019-sentence-bert}. On the other hand, Sentence-BERT \citep{reimers-2019-sentence-bert}, a language model built on top of BERT, is explicitly trained to generate sentence embeddings where semantically similar sentences are closer in the embedding space. In the next section, we explain how Sentence-BERT can be used to generate an embedding for a triple.

%% file: proposed-framework.tex
\section{The Framework}
\label{sec:framework}
We materialize the idea of using pre-trained language models for temporal knowledge graph completion into a framework called TEMT (\textbf{T}ext \textbf{E}ncoder \textbf{M}eets \textbf{T}ime). The framework that gives a plausibility score to a quadruple $\langle$s, r, o, t$\rangle$ where t is a single time point. It models a quadruple's textual and temporal information independently via text and time encoders, fuses them, and gets a plausibility score for time interval prediction. 

Figure \ref{Framework} gives an overview of the TEMT framework. Note that TEMT is different from embedding-based methods and it employs feature extraction. The text encoder takes textual information of the triple and outputs the \textit{triple embedding} $e_{sro}$. The time encoder takes temporal information and outputs the \textit{time point embedding} $e_{t}$. These two representations are fused to output a validity score. 

\subsection{Embedding Triples with a Text Encoder} 
The text encoder packs the names and descriptions of triple elements as a single sentence and returns a vector. As our text encoder, we leverage a pre-trained language model to benefit from its representation power. Our text encoder is a pre-trained Sentence-BERT \citep{reimers-2019-sentence-bert} model\footnote{The name of the model used is all-mpnet-base-v2.}. Inspired by \citep{yao_kg-bert_2019}, we form a single textual sentence $S_{sro}$ for a triple to feed Sentence-BERT.

\begin{align}
\label{Ssro}
    S_{sro}= \mathcal{N}_s + \mathcal{N}_r + \mathcal{N}_o + (\mathcal{D}_s + \mathcal{D}_o),    
\end{align}

where $S_{sro}$ is a string concatenation of the names and descriptions of the entities and relations, $\mathcal{N}$ refers to names and $\mathcal{D}$ refers to entity descriptions. The text encoder then outputs $e_{sro} \in \mathbb{R}^{\textit{d}}$ which we call triple embedding:

\begin{align}
\label{esro}
    e_{sro} = TextEncoder(S_{sro})
\end{align}

Our main motivation to leverage a language model as a text encoder is two-fold. Firstly, the language model captures the interactions between the subject, relation, and the object and outputs a semantically rich contextualized embedding of the fact. Secondly, language models can model unobserved entities and therefore support inductive reasoning. 

\subsection{Embedding Time Points with a Time Encoder}
By the independent nature of time, we claim that the representation of time information should be independent of the fact. Many previous works \citep{jain-etal-2020-temporal, 10.1145/3184558.3191639} learn the time vectors within the same space as entities and relations and use a scoring function that performs on this vector space. However, this may not allow us to capture the possible interactions between time points, for instance, consecutive years may not be modeled correctly. Hence, in this work, following previous studies \citep{DBLP:journals/corr/abs-2112-05785, 10.1145/3459637.3482416,10.1145/3336191.3371847}, we use positional encoding \citep{NIPS2017_3f5ee243} to produce vector representations of time points. In other words, the positional encoding method embeds time points into a vector space. As such, our time encoder takes a time point and returns a vector. Given some time point $t$ and a reference time point $t_{min}$, the $j$-th component of a time point embedding for \textit{t} is defined as follows:

\begin{align}
    TimeEncoder(\textit{t}, t_{min})[j] = \begin{cases}
    \sin\left(\frac{t - t_{min}}{10000^{i/d'}}\right) & \text{if}\ j=2i \\
    \cos\left(\frac{t - t_{min}}{10000^{i/d'}}\right) & \text{if}\ j=2i+1
    \end{cases}
\end{align}

where the term $t - t _{min}$ refers to the position of $t$ relative to the earliest time point $t_{min}$ in $\mathcal{T}$ and $d'$ is the dimension of the time point embedding. Intuitively, the time point embedding can be thought of as a position in time. The time encoder requires a first time point $t_{min}$ as a reference point, then the other time points will be positioned relative to this reference point. For the sake of brevity, we omit $t_{min}$ from the time encoder function and simply write as $TimeEncoder(t)$. The time encoder outputs $e_{t} \in \mathbb{R}^{\textit{d'}}$ which we call time point embedding.
\begin{equation}
       e_{t} = TimeEncoder(\textit{t})
\end{equation}
We emphasize the two properties of positional encodings: each time point corresponds to a unique vector and the vectors of close time points are closer in the vector space. This enables us to model the dependencies across different time points and the notion of ordering. Moreover, in contrast to previous work, the time encoder can represent unobserved time points. Although we do not focus on temporal inductiveness in this paper, the time encoder can potentially be used for performing predictions on future or unseen time points. 

\subsection{Fusing and Training}
\label{Fusing and Training}
\subsubsection{Fusing Triple and Time Point Embedding}
In the previous sections, we introduced two functions, namely $TextEncoder$ and $TimeEncoder$, that allow us to produce embeddings of triples and time points respectively. We are now ready to discuss how these embeddings, from different spaces, can be combined (or fused) together. 
Similar to \citep{ostendorff2019enriching, Gu2021APF}, by treating the textual and temporal features as different modalities, TEMT combines triple and time embeddings using a multi-layer perceptron (MLP). The fusion of these two embeddings produces a time-aware representation of a quadruple. Formally, given a quadruple $q$, the time-aware representation $e_{q}$ is obtained as follows:  

\begin{align}
 \label{eq}
 e_{q} = (W_1 v_{concat} +b_1) \in \mathbb{R}^{{k}} 
\\
v_{concat} = [e_{sro} ; e_{t}]\in \mathbb{R}^{d+d'} \nonumber 
\end{align}

where [ ; ] denotes concatenation operation, $W_1 \in \mathbb{R}^{{k x (d+d')}}$ and $b_1 \in \mathbb{R}^{{k}}$ denote the learnable parameters and k is a new dimension and $k < (d+d')$.

The alternative option to fusing would be adding the time point to the triple sentence $S_{sro}$ in Equation (\ref{Ssro}) and feeding the language model with this sentence. However, it is shown that pre-trained language models are not good at number representations \citep{DBLP:journals/corr/abs-2002-12327}. Our preliminary analysis also demonstrated that language models are not good at temporal reasoning such as ordering events and interval arithmetic. In addition, they are not robust to small perturbations in time information. 

\subsubsection{Quadruple Scoring Function}
Although most methods use a fixed distance function for scoring triples or quadruples, there are some methods such as ConvE \citep{dettmers2018conve} and ConvKB \citep{article} that learn the parameters for a scoring function. Similarly, we employ a parametric scoring function to output a plausibility score for a quadruple of a given TKG:
\begin{equation}
\label{scoring}
    f(s, r, o, t) = W_2e_{q} + b_2
\end{equation}
where $W_2 \in \mathbb{R}^{\textit{1 x k}}$ and $b_2 \in \mathbb{R}$  are  learnable parameters of the final layer of the neural network. Before feeding the input to this final layer, we use ReLU \citep{agarap2019deep} as an activation function.

\subsubsection{Negative Sampling}
\label{sec:neg-sample}
The model learns by distinguishing valid quadruples from incorrect quadruples. To this end, TEMT employs two different types of negative sampling. The first type is called \textit{entity-corrupted negative sampling}. In this approach, the set of negative quadruples $D^-_{\langle s,r,o, t\rangle}$ is created by corrupting the subject or the object of a given quadruple ${\langle s,r,o, t\rangle}$ as shown below:
\begin{align}
D^-_{\langle s,r,o, t\rangle} = \{\langle s', r, o, t \rangle \not\in D^+ | s' \in \mathcal{E}\} \cup \{\langle s, r, o', t \rangle \not\in D^+ | o'\in \mathcal{E}\}. \nonumber
\end{align}
where $D^+ = \mathcal{Q}_{train}$ denotes the set of positive quadruples.

The second one is called \textit{time-corrupted negative sampling} \citep{Cai_2021}. In this approach, the set of negative quadruples $D^-_{\langle s,r,o, t\rangle}$ is created by corrupting the time point of a given quadruple ${\langle s,r,o, t\rangle}$ as the following: 

\begin{center}
        \begin{math}
         D^-_{\langle s,r,o, t\rangle} = \{\langle s, r, o, t' \rangle \not\in D^+ | t' \in \mathcal{T}\}.  
        \end{math}
\end{center}
   
The way of sampling time-corrupted negatives depends on the time category of the positive quadruple: $t^{\prime}<t_{start}$ if right-open interval, $t^{\prime}>t_{end}$ if left-open interval, and $t^{\prime} \notin [t_{start}, t_{end}]$ if closed interval. 

\subsubsection{Training}
Similar to \citep{NIPS2013_1cecc7a7}, we use the following margin-based ranking loss for training:
\begin{equation}
\operatorname{\mathcal{L}} = \sum_{q_p \in D^+} \sum_{q_n \in D^-_{q_p}} \max (0,f(q_n) - f(q_p) + \gamma)
\end{equation}

\noindent
where $q_p$ is a positive quadruple,  $q_n$ is a negative quadruple, $\gamma$ is the margin value and f is the scoring function from Equation (\ref{scoring}). The model is trained to give higher scores to positive quadruples (with a given margin $\gamma$) than negative quadruples.

%% file: experiments.tex
\section{Experiments}
\label{sec:experiments}
\subsection{The Datasets}
We perform our experiments on two interval-based TKGs: YAGO11k and Wikidata12k \citep{dasgupta-etal-2018-hyte}. 
YAGO11k is created from YAGO3 knowledge graph \citep{Mahdisoltani2015YAGO3AK} and Wikidata12k is a subgraph of a preprocessed version of Wikidata \citep{10.1145/3184558.3191639}. In both datasets, each fact has a time interval attached to it and each entity has at least two edges. 

Furthermore, ind-YAGO11k datasets and ind-Wikidata12k are inductive splits that we generate from YAGO11k and  Wikidata12k. The split process is discussed in Section \ref{inductive-dataset-creation}. The details of the four datasets are given in Table \ref{tab:dataset-statistics}. We enhance the datasets with the names and descriptions of entities and relations. The next section provides the data pre-processing details.

\begin{table}[t!]
\centering
    \caption{Dataset Statistics}
    \begin{tabular}{lrrrrr}
        \toprule
        Dataset  & Entity & Relation & Train  & Valid  &  Test \\
        \midrule
           YAGO11k  & 10,623 & 10 & 16,408 & 2,050 & 2,051 \\
   Wikidata12k & 12,554 & 24 & 32,497 & 4,062 & 4,062 \\ 
       ind-YAGO11k  & 10,623 & 10 & 12,330 & 3,726 & 4,453 \\
    ind-Wikidata12k  & 12,554 & 24 & 27,330 & 6,354 & 6,937 \\
        \bottomrule
    \end{tabular}
    \label{tab:dataset-statistics}
\end{table}

\subsubsection{Dataset Pre-processing}

For Wikidata12k, the entity names and descriptions are taken from their corresponding Wikipedia pages. For YAGO11k, the entity and relation names are already available in the dataset. Similar to Wikidata12k, we extract the entity descriptions for YAGO11k from Wikipedia pages. For both datasets, the entity descriptions are limited to one sentence.

We fix the time granularity as "year" and drop the months and days similar to  \citep{jain-etal-2020-temporal, Cai_2021}. For each quadruple in the training set that has closed-interval, i.e., $\langle$s, r, o, [$t_{start}$, $t_{end}$]$\rangle$, we get two training data points $\langle$s, r, o, $t_{start}$$\rangle$ and $\langle$s, r, o, $t_{end}$$\rangle$. An alternative would be to get all intermediate time points between $t_{start}$ and $t_{end}$. However, this approach would result in over-sampling for long relations. Lastly, for the cases where either $t_{start}$ and $t_{end}$ is unknown, we only consider the known time point.
 
\subsubsection{Dataset Preparation for Inductive Reasoning}
\label{inductive-dataset-creation}
To test TEMT's ability to generalize on unseen entities, we design new splits based on YAGO11k and Wikidata12k and refer to them as ind-YAGO11k and ind-Wikidata12k, respectively. For inductive reasoning, the validation and test sets should have some entities that are not in the training set. We employ the algorithm from \citep{daza_inductive_2021} to create the new splits. The algorithm samples an entity and removes this entity from the graph $\mathcal{G}$ if this removal does not result in any isolated node or any relation type with less than 100 edges in the graph. The removed entity and its edges are then added either to the validation set and or to the test set. Thus, each triple in the test set has either a new subject or a new object. The test set has 1062 and 1255 unseen entities for YAGO11k and Wikidata12k, respectively.

The algorithm works in triple level therefore assumes that the underlying graph is static by ignoring the validity time intervals. Therefore, each split has different triples, not quadruples. As a last step, we attach the corresponding time information to each triple to convert it to a quadruple. 

\subsection{Inference and Evaluation Metrics}
\label{eval-metrics}
Since TEMT is designed to score quadruples with time points, not with time intervals, we obtain a score for each time point (in our case for each year) in the test set during the inference time. Then we need to aggregate the scores for each time point to output a time interval. We turn these scores into probabilities using the softmax function. Given a quadruple $\langle s, r, o, t\rangle$, its probability is computed as: 
\begin{align}
P(t|s,r,o) = {\mathit{softmax}(f({\langle s,r,o, t\rangle}))= \frac{\mathit{exp}(f({\langle s,r,o, t\rangle})}{\sum_{t' \in \mathcal{T}} \mathit{exp}(f({\langle s,r,o, t'\rangle}))}} \nonumber
\end{align}
Then we use greedy-coalescing algorithm \citep{Cai_2021} that takes the list of probabilities for each year and outputs $k$ most possible time intervals.

\begin{figure}[t!]
    \centering
    \captionsetup{justification=centering,margin=1cm}
    \includegraphics[width=0.45\textwidth]{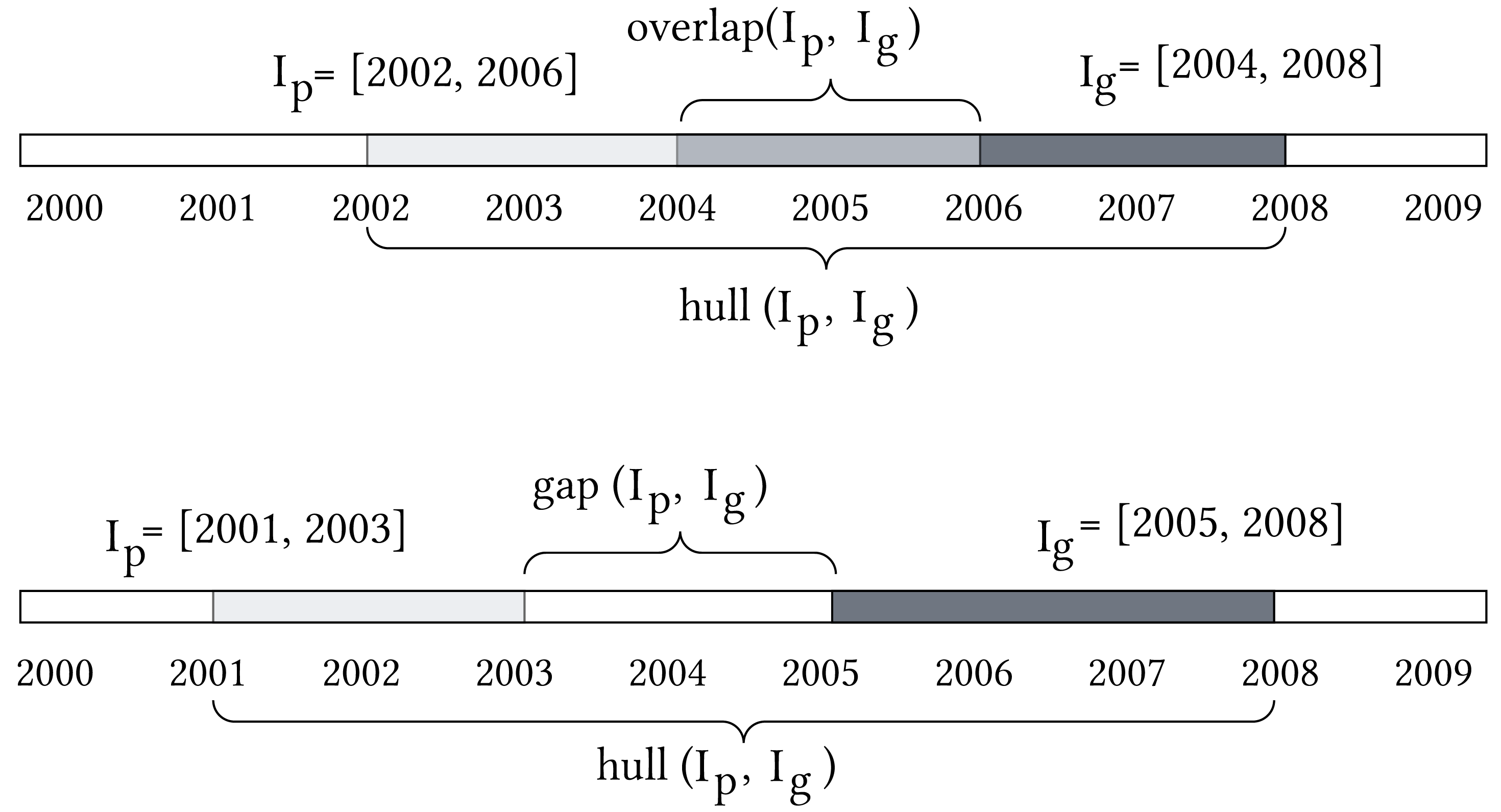}
        \caption{Time Prediction Evaluation Terms}
    \label{intervals} 
\end{figure}
    
For time interval prediction, we use some interval metrics that compare the predicted interval $I_p = [t_{start}^p, t_{end}^p]$ and the ground-truth interval $I_g=[t_{start}^g, t_{end}^g]$. Ideally, $I_p$ is completely the same as $I_g$ or they have some overlap. If there is no overlap, at least $I_p$ and $I_g$ should be close to each other. We use the following interval metrics that take these into account.

The first metric is called gIOU (generalized intersection over union)\citep{8953982} and is defined as follows:

\begin{align}
    gIOU \left(I_p, I_g\right)=  IOU \left(I_p, I_g\right) - \frac{|gap(I_p, I_g)|}{|hull(I_p, I_g)|} 
\end{align}

where 
\begin{equation}
    IOU \left(I_p, I_g\right)= \frac{|overlap(I_p, I_g)|}{|I_p|+|I_g|-|overlap(I_p, I_g)|} \nonumber 
\end{equation}
 $gap(I_p,I_g)$ is the gap between $I_p$ and $I_g$ and the hull is the shortest interval that covers both $I_p$ and $I_g$.  More formally, referring to $I_p=[t_{start}^p, t_{end}^p], I_g=[t_{start}^g, t_{end}^g]$, these functions can be defined as follows:

\begin{align*}
 gap([t_{start}^p, t_{end}^p],[t_{start}^g, t_{end}^g])&~ = [t_{end}^p, t_{start}^g], \\    hull([t_{start}^p, t_{end}^p],[t_{start}^g, t_{end}^g]) &~= [t_{start}^p, t_{end}^g]\text{ and }\\
 overlap([t_{start}^p, t_{end}^p], [t_{start}^g, t_{end}^g]) &~= [t_{start}^g, t_{end}^p] \text{.}
\end{align*}

Similarly $gap(I_g, I_p)$, $overlap(I_g, I_p)$ and $hull(I_g, I_p)$ can be defined and the length of an interval is defined as $|[t_{start},t_{end}]|=t_{end}-t_{start}+1$. As an example, Figure \ref{intervals} illustrates the terms used in the metrics. We demonstrate two scenarios when the predicted interval overlaps  with the gold interval (top figure) and when it does not (bottom figure). In the case of the former, given $I_p=[2002, 2006]$
and $I_g=[2004, 2008]$, we get the following:  
the $hull(I_p, I_g)=[2002, 2008]$, and the $overlap(I_p, I_g)=[2004, 2006]$. In the latter case, given $I_p=[2001, 2003]$ and $I_g=[2005, 2008]$, then the $hull(I_p, I_g)=[2001, 2008]$ and $gap(I_p, I_g)=[2003, 2005]$.

The second metric is aeIOU \citep{jain-etal-2020-temporal}, affinity enhanced intersection over union. It is defined as follows:

\begin{align}
  aeIOU \left(I_p, I_g\right)= \begin{cases}
    \frac{|overlap(I_p, I_g)|}
    {|hull(I_p, I_g)|}
    & |overlap(I_p, I_g)| >0, 
    \\ 
    \frac{1}{|hull(I_p,I_g)|}
    & \text{otherwise.}
    \end{cases}    
\end{align}

The drawback of aeIOU  that it outputs the same scores for both $I_p$ and $I_{p^*}$ if hull($I_p$, $I_g$) = hull($I_{p^*}$, $I_g$), ignoring the fact that one of them can be closer to $I_g$. In order to address this drawback, the study in \citep{Cai_2021} introduces a new metric called gaeIOU (generalized aeIOU).

\begin{align}
    gaeIOU \left(I_p, I_g\right)= \begin{cases}
    \frac{|overlap(I_p, I_g)|}
    {|hull(I_p, I_g)|}
    & |overlap(I_p, I_g)| >0,  
    \\\
    \frac {|gap(I_p, I_g)|^{-1}}{|hull(I_p, I_g)|}
    & \text {otherwise.}
    \end{cases} 
\end{align}

Using the three metrics, in Section~\ref{sec:experiments}, we report time interval prediction results based on gIOU@k, aeIOU@k, and gaeIOU@k. For each metric, @k denotes the best result out of k possible intervals that greedy-coalescing algorithm produces.

\subsection{Experimental Setup}
\label{Experimental setup}
The dimension $d$ for the triple embedding $e_{sro}$ is 768 and  the dimension $d'$ for time point embedding is 64. We set $k$, in Equation (\ref{eq}), to 64. We use 128 time-corrupted negative samples, as discussed in Section \ref{Fusing and Training}. For all experiments, we train our model with Adam  optimizer \citep{DBLP:journals/corr/KingmaB14} for 50 epochs with a learning rate of 0.001 and margin value $\gamma = 2$. For time interval prediction, we set the threshold for greedy-coalescing to 0.65. We report the effect of hyperparameters in \ref{details-hyperparameters}.

\subsubsection{Model Variants}
We use two different variants of TEMT, namely, TEMT$_{N}$ and TEMT$_{ND}$. The variant TEMT$_{N}$ is trained without the entity descriptions but only with the entity and relation names. In order to reflect this change, 
we modify  Equation (\ref{Ssro}) as \begin{math}S_{sro}= N_s + N_r + N_o\end{math}. On the other hand, the variant TEMT$_{ND}$ is trained with both entity descriptions and names. Regarding the sequence length for the language model, we keep the default setting in Sentence-BERT which is 128 tokens.

\subsubsection{Baselines}
In order to compare TEMT variants against the state of the art, we identify 4 different TKGC methods: HyTE \citep{dasgupta-etal-2018-hyte}, TNT-ComplEx \citep{lacroix_tensor_2020}, TIMEPLEX-base \citep{jain-etal-2020-temporal} and TIME2BOX-TNS \citep{Cai_2021}. TIMEPLEX has two variants: TIMEPLEX-base and TIMEPLEX. Unlike the other baselines, TIMEPLEX relies on temporal constraints to improve its performance. However, TIMEPLEX-base does not follow the same. Here, we report the results of the latter. All of the baselines are transductive and do not use pre-trained language models for learning entity and relation embeddings. To the best of our knowledge, TEMT is the only method that supports inductive reasoning for the task of time interval prediction.  Lastly, following our baselines, we only report on the test instances that contain known time points which is compatible with our evaluation metrics. This means that we do not report results on quadruples, in the test set, that contain unknown start or end time points. 

\input{tp.tex}

\begin{table}[t!]
      \caption{Triple Classification Results on Test Set.}
    \centering
    \begin{tabular}{lrrr}
        \toprule
        Dataset & Accuracy (\%) \\
        \midrule
    YAGO11k &  89.12\\
   Wikidata12k & 91.55\\ 
   ind-YAGO11k  & 88.64  \\
   ind-Wikidata12k & 89.82   \\
        \bottomrule
    \end{tabular}
    \label{tab:triple-classification}
\end{table}

  \begin{table*}[t!]
    \centering
        \captionsetup{justification=centering,margin=1cm}
    \caption{Best gaeIOU@1 results with different negative sample types.}
    
    \begin{tabular}{lrrrr}
        \toprule
        Negative Sample Type & YAGO11k  & Wikidata12k & ind-YAGO11k & ind-Wikidata12k  \\
        \midrule
   Entity-corrupted &   2.49 & 8.63 &  2.22 & 7.51  \\  
   Time-corrupted  &  9.77 & 12.58  & 12.19 & 11.21    \\
        \bottomrule
    \end{tabular}
    \label{tab:negative-sample-type}
    
\end{table*}
    \begin{table*}[t!]
    \captionsetup{justification=centering,margin=1cm}
        \caption{Example predictions from Yago11k and Wikidata12k datasets. The descriptions of entities are not displayed although utilized in the inference.}
        \centering
        \begin{tabular}{lrrrrr}
            \toprule
            Triple & Gold answer & 1$^{st}$ prediction & 2$^{nd}$ prediction & 3$^{rd}$ prediction  \\
            \midrule
            \textit{Kaká member of sports team Hertha BSC} & \textbf{[2008, 2012]} & [2008, 2014] & [2011, 2012] & [2004, 2011] \\
            \textit{Ippling located in the administrative territorial entity Bezirk Lothringen} & \textbf{[1871, 1920]} & [1860, 1920] & [1871, 2018] & [1919, 1920]\\
            \textit{Paulo Lopes (footballer) plays for S.L. Benfica} & \textbf{[1997, 2002]} & [1997, 2004] &  [1999, 2000] &  [2000, 2008]\\
            \textit{Jeff Morrow is married to Anna Karen Morrow} & \textbf{[1947, 1993]} &[1956, 1992]&[1959, 1960]&[1960, 2013]\\
            \textit{Henry Clay is affiliated to  Whig Party (United States)} &\textbf{[1833, 1852]} & [1830, 1862] & [1817, 1847] & [1818, 1829] \\

            \bottomrule
        \end{tabular}            
        \label{tab:illustrative}
\end{table*}

\input{ind.tex}
\input{qualitative}

%% file: tp.tex
\begin{table*}[t!]
  \label{tab:commands}          \captionsetup{justification=centering,margin=1cm}
    \caption{Transductive Time Interval Prediction Experiment Results on YAGO11k and Wikidata12k Datasets. Results marked (*) are taken from ~\cite{Cai_2021}, results marked (†) are reproduced by us, and the others are taken from  ~\cite{jain-etal-2020-temporal}. "-" denotes unavailable results.}
  \begin{tabular}{ccccccl}
    \toprule
 Methods & gIOU@1   & aeIOU@1 &  gaeIOU@1 & gIOU@10 & aeIOU@10 &  gaeIOU@10  \\ 
 \midrule
             &  & & YAGO11k & & &\\
 \midrule
     HyTE & 15.96 & 5.41 & - & - & - & - \\
    TNT-ComplEx & 20.78 & 8.40 & - & - & - & -\\
    TIMEPLEX-base† & 23.77 & 12.62 & 6.92 & 48.30 & \textbf{34.63} & 26.63 \\
    \midrule
    TEMT$_{N}$ &  \textbf{39.85} & 13.05 & \textbf{10.05} &  58.78 & 32.89 & 29.25 \\
    TEMT$_{ND}$ &  38.60 & \textbf{13.48} & 9.61 & \textbf{60.65} & 34.33 & \textbf{30.34}  \\
     \midrule

                 &  & & Wikidata12k & & &\\
 \midrule
    HyTE & 14.55 & 5.41 & - & - & - & - \\
    TNT-ComplEx & 36.63 & 23.35 & - & - & - & - \\
    TIMEPLEX-base† & 39.44  & \textbf{26.14} & 17.23 & 69.00 & 46.82 & 42.98 \\
    TIME2BOX-TNS* & 42.30 & 25.78 & \textbf{17.41} & \textbf{70.16} & \textbf{50.04} & \textbf{47.54} \\
    \midrule
    TEMT$_{N}$ &  39.35 & 12.90 & 8.81 & 61.68 & 34.97 & 30.71 \\
    TEMT$_{ND}$ & \textbf{43.52} & 17.13 & 12.58 & 65.84 & 42.00 & 38.43  \\
    
    \bottomrule
  \end{tabular}
  \label{tab:time-prediction}
\end{table*}

\begin{table}[t!]
  \label{tab:commands}
    \captionsetup{justification=centering,margin=1cm}
        \caption{Inductive Time Interval Prediction Experiment Results on ind-YAGO11k and ind-Wikidata12k Datasets.}
\setlength{\tabcolsep}{1pt}.
  \begin{tabular}{ccccccl}
    \toprule
 Methods & \begin{turn}{60} gIOU@1\end{turn}   & \begin{turn}{60}aeIOU@1\end{turn}  &  \begin{turn}{60}gaeIOU@1\end{turn}  & \begin{turn}{60}gIOU@10\end{turn}  & \begin{turn}{60}aeIOU@10\end{turn}  &  \begin{turn}{60}gaeIOU@10 \end{turn} \\

 \midrule
              &  & &  ind-YAGO11k & & &\\            
 \midrule

TEMT$_{N}$ & \textbf{39.07} & 14.23 & \textbf{10.32} & \textbf{61.53} & 35.90 & 32.79  \\
TEMT$_{ND}$ &37.20 & \textbf{14.81} & 10.15 & 60.07 &\textbf{36.73} & \textbf{33.32} \\
 \midrule

             &  & & ind-Wikidata12k & & &\\
 \midrule
        TEMT$_{N}$ & \textbf{39.78} & 12.94 & 9.18 & 60.88 & 34.92 & 31.07\\
        TEMT$_{ND}$ & 38.43 & \textbf{16.43} & \textbf{11.01} & \textbf{64.50} & \textbf{40.06} & \textbf{36.63} \\

            \bottomrule

  \end{tabular}
  \label{tab:ind-time-prediction}
\end{table}

\subsection{Transductive Time Interval Prediction}
\label{timeintervalprediction}
In this experiment, the task is to predict the validity time intervals of facts in TKGs, namely predicting $\langle$s, r, o, ?$\rangle$. We compare the TEMT variants with the baselines and report the transductive time interval prediction results for YAGO11k and Wikidata12k in Table \ref{tab:time-prediction}. For Wikidata12k, we observe that TEMT variants show improvements in the gIOU@1 metric in comparison with the baselines. For aeIOU@k and gaeIOU@k, which are more stringent metrics as discussed in \ref{eval-metrics}, TEMT variants are outperformed by  the baselines. 

Assume that two facts have the same subject, relation, object, and have very close time points. Since their triple embeddings are the same vectors and their time point embeddings are very close to each other, TEMT outputs very similar scores and this may harm TEMT's performance. However, TEMT is competitive with the state-of-the-art on the metrics gIOU@10, aeIOU@10 and gaeIOU@10.
Note that the results of TIME2BOX-TNS are taken from the paper~\cite{Cai_2021}. We could not reproduce the results for Wikidata12k and test the method on YAGO11k as neither the source code nor the details for pre-processing the datasets is available. In addition, TIME2BOX-TNS does not provide results for the YAGO11k dataset. 

On the YAGO11k dataset, TEMT outperforms the baselines in all the metrics but aeIOU@10. Notably, in the gIOU@1 metric, TEMT achieves ~16\% percentage points more than the next best competitor TIME2BOX-TNS. 

Comparing the variants, we observe that TEMT$_{ND}$ performs better than TEMT$_{N}$ in most cases. This observation supports the claim that entity descriptions improve the context and, therefore, help to create more meaningful semantic triple embeddings. We also observe that TEMT variants are better at capturing the start and the end years compared to intermediate years, which possibly hurts the time interval prediction performance. The possible explanation is that the text corpora that the language model is trained on generally contain either the starting date or end date. In addition, the textual descriptions of entities may contain temporally irrelevant descriptions. For instance, for time point 2000, we may get an entity description from Wikipedia that is updated in 2020.

%% file: ind.tex
\subsection{Inductive Time Interval Prediction}
In this experiment, we perform inductive time interval prediction on newly generated inductive datasets ind-YAGO11k and ind-Wikidata12k. Since all our baselines support only transductive reasoning, they cannot be compared with TEMT. Hence, we exclude them from this experiment. 

The inductive time prediction results are reported in Table \ref{tab:ind-time-prediction}. The results show that TEMT's performance on inductive datasets is very close to the transductive setting (Table~\ref{tab:time-prediction}). This shows the generalization power of TEMT on unseen entities by the usage of pre-trained language models given the fact that ind-YAGO11k and ind-Wikidata12k have 1062 and 1255 unseen entities in test set, respectively. Another observation is that we do not see any significant drop in performance although the models are trained on $\sim$ 4,000 fewer training points than YAGO11k and $\sim$ 5,000 fewer training points than Wikidata12k. 

Moreover, similar to Section \ref{timeintervalprediction}, TEMT$_{ND}$ variant performs better in most cases. This indicates that the context that entity descriptions provide helps the model to capture the semantics of a triple better. Note that TEMT is also applicable to a fully inductive setting where there is no overlap between train and test set entities, which we leave as future work.

%% file: qualitative.tex
\subsection{Fine-grained Analysis}

\subsubsection{Triple Classification}
We investigate the representation power of triple embeddings by performing triple classification experiment. The motivation is to make sure that our text space is also meaningful like our time space. With this experiment, we predict whether a triple is correct or not. To this end, we train an MLP classifier \citep{scikit-learn} with triple embeddings ($e_{sro}$). \ref{tc-details} includes the details of the datasets created and the training settings for this experiment.
 
We report the triple classification results in Table \ref{tab:triple-classification}. The results illustrate the effectiveness of the text encoder thus support the claim that the triple embeddings are semantically meaningful and  potentially capture the structural information. Moreover, it also indicates that TEMT can handle atemporal statements, which are common in real-world knowledge graphs.

\subsubsection{The Effect of Negative Sampling}
In Table~\ref{tab:negative-sample-type}, we compare the time prediction performance of TEMT$_{ND}$ variant on two negative sampling approaches discussed in Section~\ref{Fusing and Training}. The results show that time-corrupted negative sampling strategy is more suitable for our problem. A further analysis on the effect of the number of negative samples can be found in Appendix \ref{number-of-negs}.

\subsubsection{Time Prediction Diagnosis}
Table \ref{tab:illustrative} illustrates some examples for time interval prediction experiment on both YAGO11k and Wikidata12k datasets. “Triple” column represents some triples from the test set and the “Gold answer” column represents their correct validity time interval. The table covers the triples that occurred in different centuries and that have varying durations. The next columns report the TEMT$_{ND}$ predictions for the corresponding triple. 

We showcase that TEMT$_{ND}$ is able to output the intervals that are close to ground-truth intervals. For instance, in the first row, TEMT successfully predicts the starting point but output a longer interval than the ground-truth.  In the second row, the ending time point is predicted correctly with an earlier starting point from the gold answer. Another observation is that the predictions are mostly a subset of the gold interval so it shows that the textual information helps to predict the time period of facts.

%% file: related-work.tex
\section{Related Work}\label{related-work}
\subsection{Knowledge Graph Completion}
\label{kgc}
There have been numerous studies on static knowledge graph completion methods \citep{Ji_2022}. These methods can be roughly divided into two: knowledge graph embedding  (KGE) methods and text-based methods. Knowledge graph embedding methods represent  entities and  relations with low-dimensional vectors. They can be broadly classified into three different types: translational \citep{NIPS2013_1cecc7a7}, semantic matching \citep{https://doi.org/10.48550/arxiv.1412.6575}, and deep learning \citep{dettmers2018conve} methods. A common approach for KGE methods is to learn a function to score the plausibility of a triple. These methods perform well on many downstream tasks such as link prediction. However, they only utilize the structure of a graph and cannot easily be adapted to use additional information such as the textual  descriptions of entities and relations. Text-based methods utilize textual information in knowledge graphs to infer missing links and we discuss them in Section \ref{text-based}.

\subsection{Temporal Knowledge Graph Completion}
Although the majority of prior research has focused on static KGs, there has been a growing interest in exploring evolving knowledge graphs \citep{DBLP:journals/corr/abs-2201-08236}. In this section, we discuss interval-based TKG completion methods. A common approach for these methods is to incorporate time into the scoring functions of static KGE methods. For instance, HyTE \citep{dasgupta-etal-2018-hyte} learns to assign a hyperplane for each time point, which can be interpreted as a static snapshot of the TKG. It learns the temporal embeddings of entities and relations for each hyperplane by applying a translational scoring function from TransE \citep{NIPS2013_1cecc7a7}. Since the hyperplanes for the time points are learned independently, HyTE is not able to model the dependencies between them. Moreover, both TNT-ComplEx \citep{lacroix_tensor_2020} and TIMEPLEX \citep{jain-etal-2020-temporal} are based on ComplEx \citep{trouillon2016complex} and learn complex-valued embeddings for entity, relation and time instants. TNT-ComplEx extends ComplEx by adding  a new factor and solve a tensor completion problem. On the other hand, TIMEPLEX adds multiple time-dependent components to the scoring function and also takes into account additional learned features such as temporal constraints. TIME2BOX \citep{Cai_2021} extends the idea of box embeddings \citep{ren2020query2box} by introducing time-aware boxes and allows both atemporal and temporal facts. Unlike TEMT, these models do not benefit from external information such as textual descriptions of entity and relations. Furthermore, these models are transductive so they cannot predict on unseen entities.

\subsection{Text-enhanced (Temporal) Knowledge Graph Completion} 
\label{text-based}
In this section, we discuss text-based methods introduced in \ref{kgc}. Text-based methods incorporates the names and the descriptions of entities and relations (namely text-enhanced knowledge graphs). Some recent works on text-enhanced static knowledge graphs employ pre-trained language models for static KGC \citep{yao_kg-bert_2019, daza_inductive_2021, li_siamese_2021, wang_simkgc_2022, alam_language_2022}. The textual descriptions of entities and relations are fed into language models, that store real-world knowledge in their parameters, to obtain a rich contextual representation of the entities (or relations). However, these text-based methods do not model the dynamics in which the relations between entities hold in a time interval. 

Only a few works focus on combining language models and temporal knowledge graphs. As an example, ECOLA \citep{https://doi.org/10.48550/arxiv.2203.09590} jointly optimizes an objective function for language model and temporal knowledge graph embedding by combining the loss functions. In contrast to TEMT, which extracts entity/relation names and descriptions from Wikipedia pages, ECOLA retrieves textual information from news articles that correspond to specific dates. Additionally, ECOLA does not focus on time interval prediction. Moreover, similar to TEMT, a recent method called SST-BERT \citep{chen2023incorporating} combines the semantics of entities/relations and temporal aspects to output a plausibility score. However, this model utilizes relation paths with a primary focus on relation prediction whereas TEMT focuses on time prediction.

%% file: conclusion.tex
\section{Conclusion}
In this work, we propose TEMT, a model for text-enhanced temporal knowledge graph completion. TEMT captures semantic relationships and temporal dependencies between facts. Our empirical evaluation in both transductive and inductive time interval prediction tasks demonstrates the effectiveness of TEMT.  TEMT outperforms state-of-the-art methods on the YAGO11k dataset and achieves competitive results on the Wikidata12k dataset.  In particular, to the best of our knowledge, TEMT is the first method that is capable of performing inductive time interval prediction. 

As a future work,  we plan to investigate other pre-trained language models (such as RoBERTa~\cite{liu2019roberta}),  and time encoding methods. In addition, we also plan to incorporate structural information into TEMT's fusing function. 

%% file: appendix.tex
\begin{table*}
  \caption{Time prediction performance with respect to the number of negative samples.}
\begin{tabular}{ccccccl}
    \toprule
     & gIOU@1   & aeIOU@1 &  gaeIOU@1 & gIOU@10 & aeIOU@10 &  gaeIOU@10  \\ 
    \midrule
        &  & & YAGO11k & & &\\
 \midrule
 \# Entity-corrupted   \\
 \midrule
        16 & 14.67  & 1.37  & 0.34 & 42.82 & 4.73 & 2 \\ 
         32 & 21.29 & 0.65  & 0.24 & 44.72 & 3.14 & 1.78\\
         64 & 22.61 & 5.35 & 2.49 & 39.29 & 11.55 & 7.34 \\
         128 & 4.77  & 0.46 &  0.1 & 35.09 & 1.52 & 0.68 \\   
 \midrule
 \#  Time-corrupted  \\
 \midrule
        16 & 44.24 & 11.36  & 9.05 & 58.71 & 32.45 & 28.38  \\ 
         32 & 41.17 & 12.78 & 9.77 & 59.54 & 33.38 & 29.39 \\
         64 & 39.50 & 13.14 & 9.59 & 60 & 34.02 & 30.03\\
         128 & 38.60 & 13.48 & 9.61 & 60.65 & 34.33 & 30.34 \\
 \midrule

        &  & &  ind-YAGO11k & & &\\
 \midrule
 \# Entity-corrupted \\
 \midrule
        16 & 26.96  & 4.8  & 2.22 & 40.52 & 10.81 & 6.69 \\ 
         32 & 32.75 & 1.96 & 1.08 & 47.55 & 6.99 & 4.73 \\
         64 & 11.99  & 1.29 & 0.33 & 33.01 & 2.11 & 1.11\\
         128 & 3.28 & 0.22 & 0 & 20.72 & 0.39 & 0.06 \\   
   \midrule
 
     \#  Time-corrupted   \\
 \midrule
         16 & 47.47 & 14.39 & 12.19 & 61.64 & 36.23 & 33 \\ 
         32 & 41.89  & 15.31  & 11.51 & 62.08 & 37.50 & 34.17 \\
         64 & 41.79 & 14.12  & 10.71 & 63.70 & 37.61 & 34.6 \\
         128 & 37.20 & 14.81 & 10.15 & 60.07 & 36.73 & 33.32 \\
\bottomrule
  \end{tabular}

    \label{ablation-negatives}
\end{table*}

\section{Appendix}
\subsection{Triple Classification Evaluation Setting}
\label{tc-details}
For the datasets, we convert the list of quadruples in the training and the test sets into triples by removing time intervals. For each training triple, we corrupt the head or tail randomly and create one negative example (that does not belong to the training set) to avoid class imbalance. We remove the test triples that exist in the training or validation set to prevent any information leakage. For each test triple, we create one negative example that does not appear in train, validation or test set. The sizes of the training sets are 32,690 for YAGO11k, 64,980 for Wikidata12k, 24,558 for ind-YAGO11k and 54,646 for ind-Wikidata12k. The sizes of the test sets are 4,100 for YAGO11k, 5,530 for Wikidata12k, 8,880 for ind-YAGO11k and 13,872 for ind-Wikidata12k. We create the sentences by following Equation (\ref{Ssro}) and extract the features using our text encoder (namely Sentence-BERT). 
We set L2 regularization term alpha = 0.05 and perform maximum 1000 iterations. We keep the default values of the MLP classifier in \citep{scikit-learn} for the other settings.

\subsection{Effect of Number of Negative Samples}
\label{number-of-negs}
We conduct an empirical study to see how sampling types discussed in \ref{sec:neg-sample} affect the performance of TEMT. We analyze different number of entity-corrupted and time-corrupted negative samples on YAGO11k and ind-YAGO11k datasets. The results are reported in Table \ref{ablation-negatives}. We performed the same experiments on Wikidata12k and ind-Wikidata12k as well, however, we do not include their results here, for the sake of brevity. 

The results in Table~\ref{ablation-negatives} show that the entity-corrupted negative sampling performs worse than the time-corrupted negative sampling for both datasets. Since the time interval prediction also requires the model to distinguish facts with different time points, this difference is expected. Moreover, in time-corrupted cases, the number of negative samples does not result in marginal changes in gaeIOU@1 metric, which is the most stringent metric. 

\subsection{Ablation Study}
\label{details-hyperparameters}
We explore the effect of various hyperparameters on the performance of TEMT. This also allows us to choose the optimal parameters that are discussed in Section \ref{Experimental setup}. 
To this end, we carry out a number of experiments. The results are shown in Table \ref{tab:hyper-params}. We report gaeIOU@1 results of different hyperparameters on the validation set of Wikidata12k. Although we do not observe a significant difference among the results, the setting where d'=64, learning rate=0.001, margin=2, and threshold=0.65 gives the best results. 

\begin{table}[t!]
    \centering
        \caption{ Results of hyperparameter analysis on the Wikidata12k dataset.}        
\begin{tabular}{lr ||lr}
        \toprule
        $d'$ &  gaeIOU@1 & learning rate & gaeIOU@1  \\        
        \midrule
   32 & 11.76 & 0.001 & 11.86 \\
   64 & 11.94 & 0.002 & 11.52\\ 
   128 & 11.77 & 0.003 & 11.74\\
   256 & 11.78 & 0.01 & 11.36 \\
    \midrule
    margin &  gaeIOU@1 & threshold &  gaeIOU@1 \\
        \midrule
   1 & 12.26 & 0.4 & 10.29\\ 
   2 & 12.33 & 0.5 & 11.56\\
   5 & 11.18 & 0.65 & 11.87 \\
   7 & 10.97 & 0.7 & 11.66  \\ 
        \bottomrule
    \end{tabular}
\label{tab:hyper-params}
\end{table}